# The Analysis of Local Motion and Deformation in Image Sequences Inspired by Physical Electromagnetic Interaction

XIAODONG ZHUANG[1,2] and N. E. MASTORAKIS[1,3,4]
1. WSEAS Research Department, Agiou Ioannou Theologou 17-23, 15773, Zografou, Athens, GREECE (xzhuang@worldses.org)
2. Automation Engineering College, Qingdao University, Qingdao, 266071, CHINA
3. Department of Computer Science, Military Institutions of University Education, Hellenic Naval Academy, Terma Hatzikyriakou, 18539, Piraeus, GREECE
mastor@wseas.org   http://www.wseas.org/mastorakis
4. Technical University of Sofia, BULGARIA

*Abstract:* - In order to analyze the moving and deforming of the objects in image sequence, a novel way is presented to analyze the local changes of object edges between two related images (such as two adjacent frames in a video sequence), which is inspired by the physical electromagnetic interaction. The changes of edge between adjacent frames in sequences are analyzed by simulation of virtual current interaction, which can reflect the change of the object's position or shape. The virtual current along the main edge line is proposed based on the significant edge extraction. Then the virtual interaction between the current elements in the two related images is studied by imitating the interaction between physical current-carrying wires. The experimental results prove that the distribution of magnetic forces on the current elements in one image applied by the other can reflect the local change of edge lines from one image to the other, which is important in further analysis.

## 1 Introduction

Image sequence processing is one of the main topics of digital image processing, which has significant value in theory and practical application [1-4]. Many real-world applications are based on image sequence processing such as security monitoring, traffic surveillance, 3D reconstruction of medical images, etc. One trend of sequence processing is advanced intelligent analysis which aims at accurate object recognition, human gesture recognition and even behavior recognition. The demand from practical applications gives new requirements about image sequence processing techniques, which has attracted lots of research efforts.

The change between frames in image sequence is important in automatic analysis, because the changes are usually caused by the motion or deformation of objects. Such local changes between frames can be the basis for further analysis. Here it is supposed that the frame rate in video capture is fast enough for continuous video recording and dramatic changes between adjacent frames do not exist, which is a common case nowadays. Simple change detection between frames can be achieved by frame differencing, but further motion analysis always involves local matching which is a time-consuming task with lots of computation.

It has been analyzed that the physical electro-static or magneto-static field has a feature of local-global balance, which is suitable for local feature analysis in images [5]. In recent years, physical field inspired methods have become a novel direction in image processing. Electro-magnetic field inspired method is one category of such methods, and promising results have been obtained in edge detection, corner detection, shape skeletonization, ear recognition, etc [5-13]. However, most methods in previous research are mainly inspired by electro-static field, but the analysis and imitation of magnetic field is much less. Moreover, almost all previous research concentrated on the imitation of static fields, but the dynamic interaction between images was not investigated to the authors' knowledge. In this paper, the virtual electro-magnetic interaction is studied and applied in local change analysis for image sequence processing, which provides an alternative idea of effective and efficient local motion analysis.

Edge is an important and basic image feature,

which represent the borders of objects in images. In motion analysis, it has been proved that the accuracy of motion estimation can be well guaranteed for edge points, but the estimation becomes inaccurate in the areas with little grayscale or color variation. The motion or deforming of objects will cause local change of their edges from one frame to the next frame. In another word, the local change of the edge may represent the details of the objects' motion or deforming in image sequences. Therefore, in this paper the local change of edges between images is studied by a novel approach imitating the electro-magnetic interaction.

## 2 The Interaction between Current-carrying Wires

In physics, the electromagnetic interaction between current-carrying wires is a basic and important phenomenon, which has been extensively studied and applied. In this paper, the unique characteristic in electromagnetic interaction is exploited in image analysis.

### 2.1 The magnetic field of current-carrying wire

In physics, the magneto-static field is described by the Biot-Savart law [14-17], where the source of the magnetic field is the current of arbitrary shapes which is composed of current elements. A current element $I\vec{dl}$ is a vector representing a very small part of the whole current, whose magnitude is the arithmetic product of $I$ and $dl$ (the length of a small section of the wire). The current element has the same direction as the current flow on the wire. The magnetic induction generated by a current element $I\vec{dl}$ is as following [14-17]:

$$\vec{dB} = \frac{\mu_0}{4\pi} \cdot \frac{I\vec{dl}\times\vec{r}}{r^3} \qquad (1)$$

where $\vec{dB}$ is the magnetic induction vector at a space point. $I\vec{dl}$ is the current element. $r$ is the distance between the space point and the current element. $\vec{r}$ is the vector from the current element to the space point. The operator $\times$ represents the cross product. The direction of the magnetic field follows the "right-hand rule" [14-17]. The direction distribution of the magnetic field on the 2D plane where the current element lies is shown in Fig. 1.

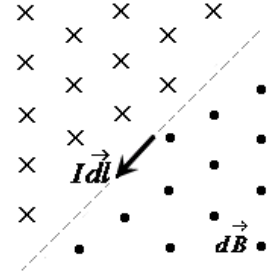

**Fig. 1 The magnetic field's direction distribution of a current element on the 2D plane**

The magnetic field generated by the current in a wire of arbitrary shape is the accumulation of the fields generated by all the current elements on the wire, which is described by the Biot-Savart law [14-17]:

$$\vec{B} = \int_D \vec{dB} = \int_D \frac{\mu_0}{4\pi} \cdot \frac{I\vec{dl}\times\vec{r}}{r^3} \qquad (2)$$

where $\vec{B}$ is the magnetic induction vector at a space point generated by the whole current of the wire. $D$ is the area where the current element exists. $\vec{dB}$ is the magnetic field generated by each current element in $D$.

### 2.2 The force on current-carrying wire by stable magnetic field

In electro-magnetic theory, the magnetic field applies force on moving charges, which is described by the Lorentz force. Derived from the Lorentz force, the magnetic force on a current element is as following [14-17]:

$$d\vec{F} = Id\vec{l}\times\vec{B} \qquad (3)$$

where $Idl$ is the current element and $dF$ is the magnetic force on it caused by the magnetic induction $B$. The direction of $dF$ satisfies the rule shown in Fig. 2. In Fig. 2, the vector of $dF$ is perpendicular to both $B$ and $Idl$.

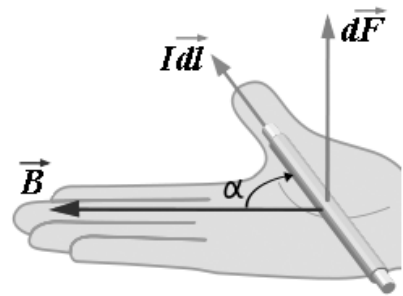

**Fig. 2 The directions of *B*, *Idl* and *dF***

A current-carrying wire of arbitrary shape consists of many current elements. The magnetic force on a wire is the summation of the force *dF* on all its current elements, which is as following [14-17]:

$$\vec{F} = \int_C I d\vec{l} \times \vec{B} \quad (4)$$

where $C$ is the integration path along the wire, and $F$ is the total magnetic force on the wire.

## 2.3 The interaction between two current-carrying wires

The current-carrying wire can generate magnetic field, meanwhile the magnetic field can put force on another wire. Thus there is interaction force between two current-carrying wires. In electro-magnetic theory, the Ampere force between two wires of arbitrary shapes is based on the line integration and combines the Biot-Savart law and Lorentz force in one equation as following [14-17]:

$$\vec{F}_{12} = \frac{\mu_0}{4\pi} \int_{C_1} \int_{C_2} \frac{I_1 d\vec{l}_1 \times (I_2 d\vec{l}_2 \times \vec{r}_{21})}{r_{21}^3} \quad (5)$$

where $F_{12}$ is the total force on wire1 due to wire2. $\mu_0$ is the magnetic constant. $I_1 dl_1$ and $I_2 dl_2$ are the current elements on wire1 and wire 2 respectively. $r_{21}$ is the vector from $I_2 dl_2$ to $I_1 dl_1$. $C_1$ and $C_2$ are the integration path along the two wires respectively.

Equation (5) is virtually the accumulation of the magnetic force on the current elements in wire1 put by the current elements in wire2. If such interaction is simulated on computers, the continuous wires should be discretized, and the integration in Equation (5) should be discretized to summation:

$$\vec{F}_d = \frac{\mu_0}{4\pi} \sum_{\vec{T}_{1j} \in C_1} \sum_{\vec{T}_{2k} \in C_2} \frac{\vec{T}_{1j} \times (\vec{T}_{2k} \times \vec{r}_{kj})}{r_{kj}^3} \quad (6)$$

where $F_d$ is the force on wire1 from wire2. Here both wire1 and wire2 are in discrete form, which consist of a set of discrete current element vectors respectively. $C_1$ and $C_2$ are the two sets of discrete current elements for wire1 and wire2 respectively. In another word, all the discrete vectors in $C_1$ constitute the discrete form of wire1, and all the discrete vectors in $C_2$ constitute the discrete form of wire2. $T_{1j}$ and $T_{2k}$ are the current element vectors in $C_1$ and $C_2$ respectively. $r_{kj}$ is the vector from $T_{2k}$ to $T_{1j}$.

Some examples of discretized current-carrying wires are shown in Fig. 3. For display in figures, the continuous current directions are discretized into 8 directions: {east, west, north, south, northeast, northwest, southeast, southwest}. Fig. 3(a) shows the discrete form of a continuous current of the rectangle shape. Fig. 3(b) shows the discrete form of a continuous current of the circle shape. Fig. 3(c) shows the discrete form of a continuous current of the straight-line shape.

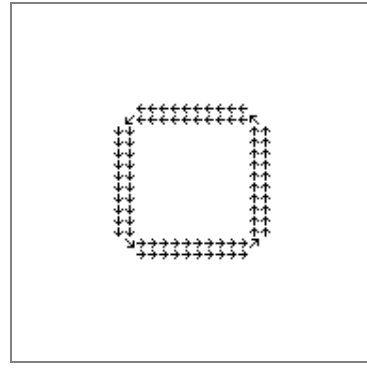

Fig. 3(a) The discrete form of a continuous rectangle current

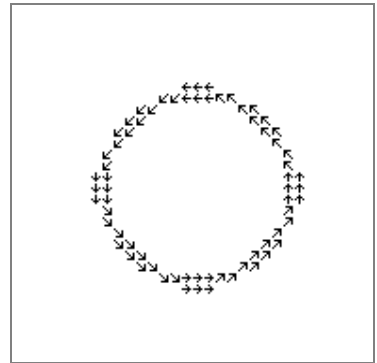

Fig. 3(b) The discrete form of a continuous circle current

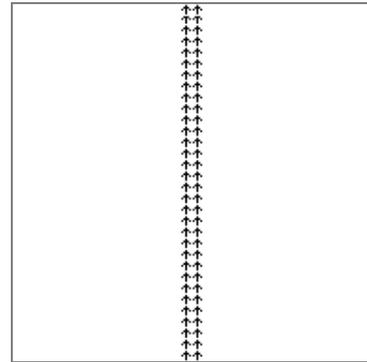

Fig. 3(c) The discrete form of a continuous straight-line current

## 3 The Virtual Currents in Images and Their Virtual Interaction

The edge feature is important and widely used in practical image processing tasks. Edges are the border of objects in an image, and the change of edge lines can reflect the moving or deforming of objects. In this paper, the virtual current along main edge lines in the image is proposed based on the significant edge extraction, and the interaction between the virtual edge currents in two images is studied.

### 3.1 Feature extraction of the "significant edge current"

In this paper, the "significant edge points" are extracted as the basis of analysis for image sequence. The significant edges are definite borders of regions, and there is sharp change of grayscale across the significant edge lines. In the proposed method, the significant edge points are first extracted in the two images to be analyzed (such as two adjacent frames in a video sequence). The virtual currents in the image are defined as the set of discrete current elements on the significant edge lines. Then the interaction force between the virtual currents in the two images is calculated, which inspires a novel approach of local change analysis in image sequences.

Canny operator is widely used to extract edge lines [18,19]. To improve computation efficiency, in this paper a simplified Canny-like method is proposed to extract the significant edges in image. First, Sobel operator is used to estimate the gradient fields of the two images to be analyzed. Then the threshold processing of the gradient magnitude is performed to reserve the definite edge points, in which only the points with a magnitude larger than the threshold value are reserved for further process. (The threshold value is set as a predefined percent of the maximum value of gradient magnitude.) After that, a "non-maximum suppression" is performed to get thin edge lines. For an edge point, it is reserved as a significant edge point only if its gradient magnitude is larger than the adjacent points on at least two of the following pairs of directions: west and east, north and south, northwest and southeast, northeast and southwest.

With the above process, the significant edge lines can be extracted. However, the direction of current element should be along the tangent direction of the wire, while the gradient vector is perpendicular to the tangent direction of the edge line. To get the virtual edge current, all the gradient vectors on the significant edge lines rotate 90 degrees. Then the vector direction after rotation is along the tangent direction of edge lines. In another word, on the significant edge lines, the virtual current elements are obtained by rotating the gradient vectors 90 degrees counterclockwise. For demonstration, a simple example is shown in Fig. 4. All the virtual current element vectors on the significant edges form the virtual currents in the image, i.e. the virtual edge current in an image is a set of discrete current element vectors on the significant edge lines.

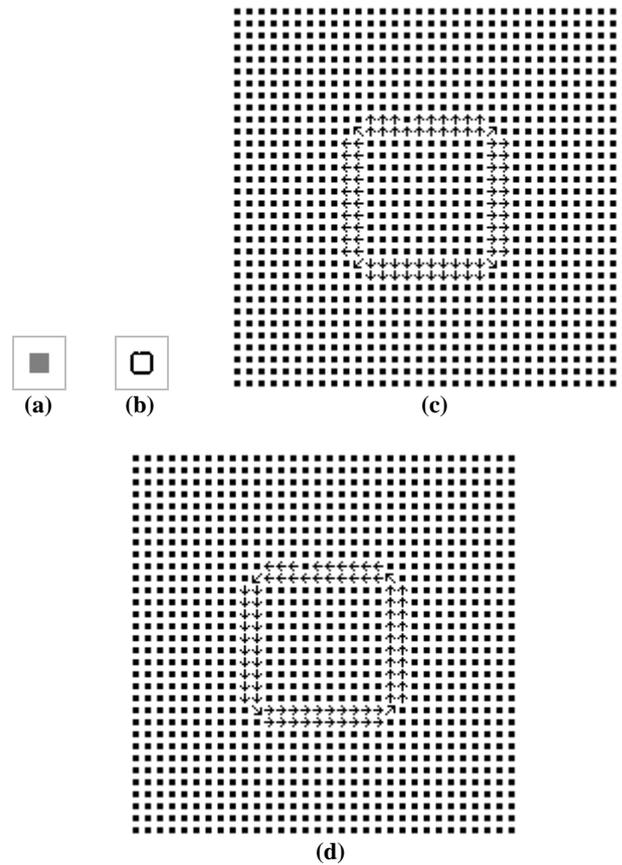

**Fig. 4 An example of the significant edge lines and the corresponding virtual current of a square shape**
**(a) The image of the square shape**
**(b) The significant edge lines of (a)**
**(c) The gradient vectors on the significant edge lines**
**(d) The discrete virtual current**

Fig. 4(a) shows a simple image of a square. Fig. 4(b) shows its significant edge lines extracted. Fig. 4(c) shows the direction distribution of the discrete gradient vectors, where the continuous gradient direction is discretized into eight directions for display: {east, west, north, south, northeast, northwest, southeast, southwest}. Fig. 4(d) shows the direction distribution of the discrete current elements, which is the rotated version of gradient vector. The dots in Fig. 4(c) and Fig. 4(d) show the points with no current elements.

Some results of virtual edge current extraction are shown from Fig. 5 to Fig. 9. Fig. 5(a) to Fig. 8(a) show some simple images of the size $32 \times 32$. Some real world images of the size $128 \times 128$ are shown in Fig. 9. The significant edge lines extracted are also shown in these results, which will be used as virtual edge currents in the following analysis.

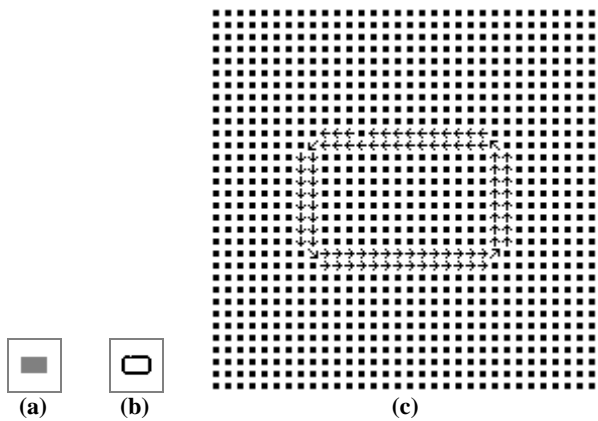

**Fig. 5 The significant edge lines and the corresponding virtual current of a rectangle shape**
**(a) The image of the rectangle shape**
**(b) The significant edge lines of (a)**
**(c) The discrete virtual current**

Fig. 5(a) shows an image of a rectangle. Fig. 5 (b) shows its significant edge lines. Fig. 5 (c) shows the virtual current elements on the significant edge lines, where the arrows show the discrete directions of discrete current elements. The dots in Fig. 5(c) show the points with no current elements.

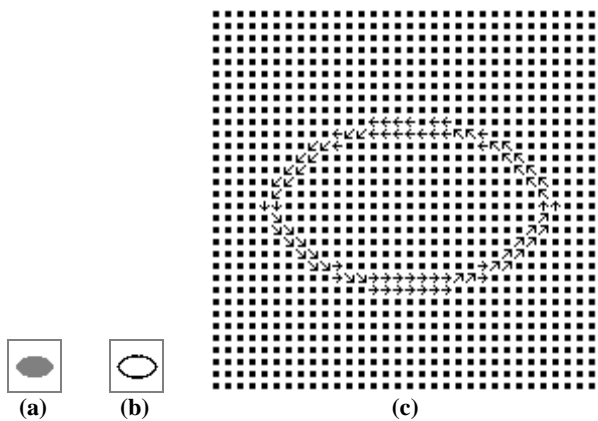

**Fig. 6 The significant edge lines and the corresponding virtual current of an ellipse shape**
**(a) The image of the ellipse shape**
**(b) The significant edge lines of (a)**
**(c) The discrete virtual current**

Fig. 6(a) shows an image of an ellipse. Fig. 6(b) shows its significant edge lines. Fig. 6(c) shows the virtual current elements on the significant edge lines, where the arrows show the discrete directions of discrete current elements.

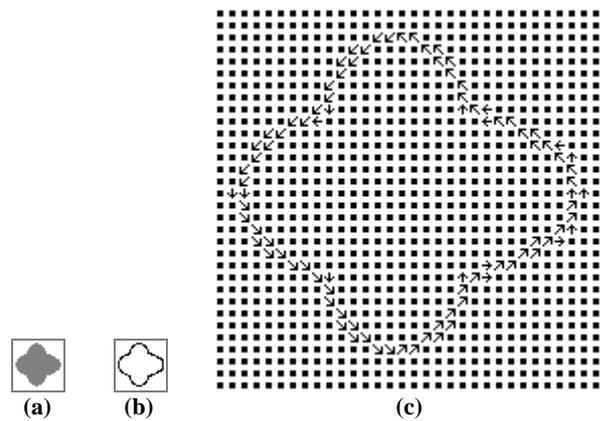

**Fig. 7 The significant edge lines and the corresponding virtual current of an irregular shape**
**(a) The image of the irregular shape**
**(b) The significant edge lines of (a)**
**(c) The discrete virtual current**

Fig. 7(a) shows an image of an irregular shape. Fig. 7(b) shows its significant edge lines. Fig. 7(c) shows the virtual current elements on the significant edge lines, where the arrows show the discrete directions of discrete current elements.

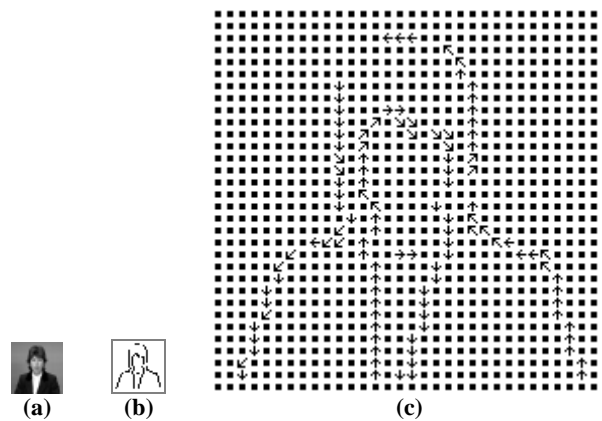

**Fig. 8 The significant edge lines and the corresponding virtual current of a shrunken image of broadcaster**
**(a) The shrunken broadcaster image**
**(b) The significant edge lines of (a)**
**(c) The discrete virtual current**

Fig. 8(a) shows the shrunken version of a broadcaster image. Fig. 8(b) shows its significant edge lines. Fig. 8(c) shows the virtual current elements on the significant edge lines, where the arrows show the discrete directions of discrete current elements.

In Fig. 9, the real world images and their significant edge lines are shown, including the images of the broadcaster, the cameraman, the peppers, the locomotive, the boat, the house and a medical image of brain.

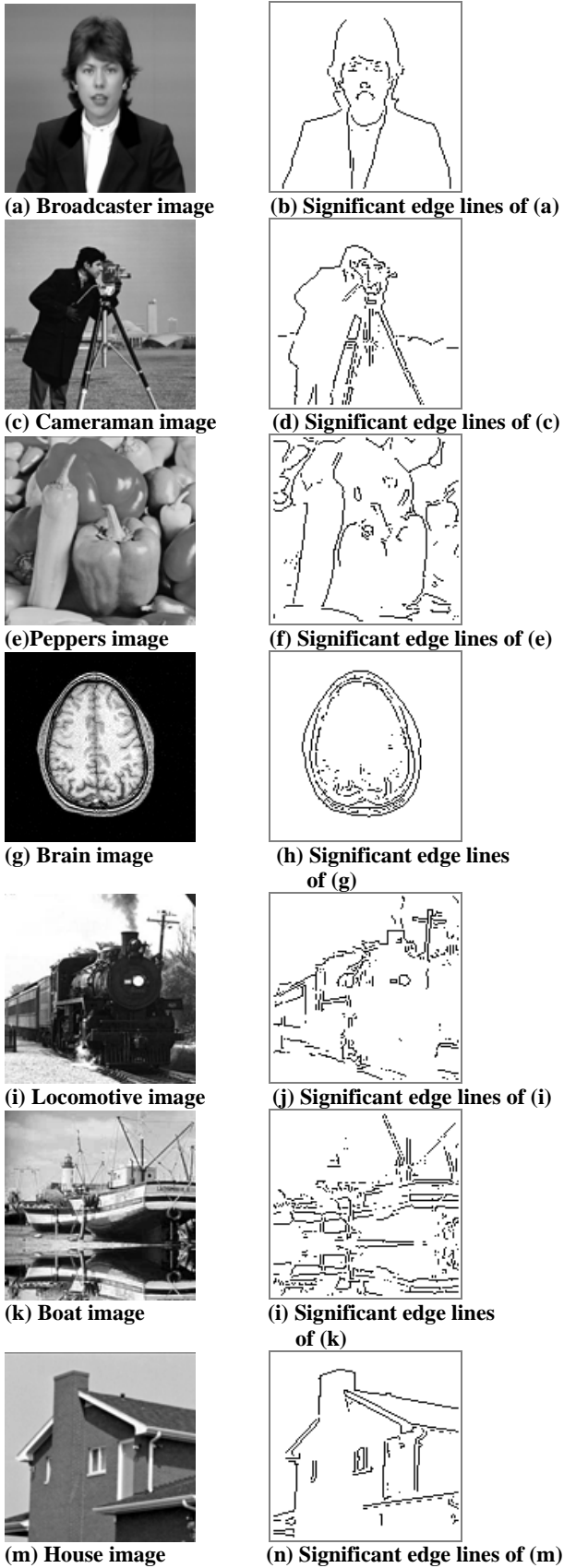

(a) Broadcaster image (b) Significant edge lines of (a)
(c) Cameraman image (d) Significant edge lines of (c)
(e) Peppers image (f) Significant edge lines of (e)
(g) Brain image (h) Significant edge lines of (g)
(i) Locomotive image (j) Significant edge lines of (i)
(k) Boat image (l) Significant edge lines of (k)
(m) House image (n) Significant edge lines of (m)

**Fig. 9 A group of real world images and their significant edge lines**

## 3.2 The interaction of the virtual edge currents between two images

If the virtual currents in the two images are extracted respectively, the interaction force between them can be calculated according to Equation (6). Suppose the sets of discrete current elements in image1 and image2 are $C_1$ and $C_2$ respectively. Each discrete current element $T_{1j}$ in $C_1$ is applied the force by all the current elements in $C_2$:

$$\vec{F}_{1j} = A \cdot \sum_{\vec{T}_{2k} \in C_2} \frac{\vec{T}_{1j} \times (\vec{T}_{2k} \times \vec{r}_{kj})}{r_{kj}^3} \quad (7)$$

where $F_{1j}$ is the force on $T_{1j}$ from $C_2$. $T_{2k}$ is a current element vector in $C_2$. $r_{kj}$ is the vector from $T_{2k}$ to $T_{1j}$. $A$ is a predefined constant value. Some simulation examples of the force on the virtual current elements are shown in Fig. 10 and Fig. 11. Here suppose the two images are on the same plane, and they coincide in position (i.e. they have the same coordinates).

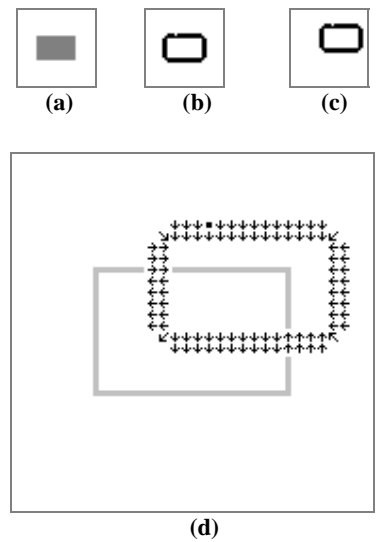

(a) (b) (c)

(d)

**Fig. 10 The virtual force on each current element in the shifted image for the rectangle image**
(a) the rectangle image
(b) the significant edge lines of (a)
(c) the significant edge lines of the shifted image
(d) the forces on the discrete current elements in the shifted image

Fig. 10(a) shows the image of a rectangle. Now move the rectangle northeast to generate another image, and the significant edge lines of the original image and the shifted one are shown in Fig. 10(b) and Fig. 10(c). The translation from (b) to (c) on *x* and *y* coordinates are 5 and -4 respectively. Fig. 10(d) shows the force direction on each current element in the shifted image, which is applied by the virtual current in the original one. The gray lines in Fig.

10(d) show the original position of the rectangle. The force on each current element in the shifted image is calculated according to Equation (7).

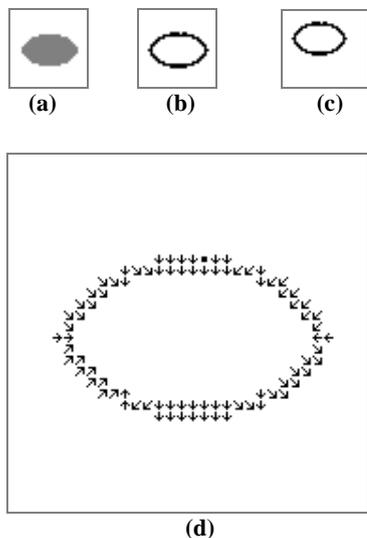

**Fig. 11 The virtual force on each current element in the shifted image for the ellipse image**
(a) the ellipse image
(b) the significant edge lines of (a)
(c) the significant edge lines of the shifted image
(d) the forces on the discrete current elements in the shifted image

Fig. 11(a) shows the image of an ellipse. Now move the ellipse northwest to generate another image, and the significant edge lines of the original image and the shifted one are shown in Fig. 11(b) and Fig. 11(c). The translation from (b) to (c) on $x$ and $y$ coordinates are -6 and -6 respectively. Fig. 11(d) shows the force direction on each current element in the shifted image, which is applied by the virtual current in the original one.

## 4 The Distribution of Virtual Interaction Force between Two Related Images

In two related images (such as two adjacent frames in a video sequence), the motion or deformation of an object will cause local changes of grayscale or color at some image positions especially at the edge points (supposing the frame rate is fast enough to capture continuously changing images). In practice, the motion and deformation of an object can occur simultaneously. Local change analysis can be the basis of motion and deformation estimation in image sequences. On the other hand, consider the factor of distance $r_{kj}$ on the denominator in Equation (7), it can be seen that in the interaction between the virtual currents, a current element in one image applies much stronger force on adjacent current elements than the remote ones in the other image. In another word, the force on a current element in $C_1$ is mainly determined by the adjacent current elements in $C_2$ (supposing image1 and image2 are on the same plane and have the same coordinates). It indicates that the virtual magnetic force on a current element in one image is mainly determined by adjacent current elements in the other image. Such local characteristic is suitable for the representation of local features (both temporal and spatial) in image sequences.

In the following experiments, deforming of object is studied as a typical case of local change in images. Deforming is an important type of transformations between images, which has been widely studied in practical applications [20,21]. In the following experiments, the virtual electro-magnetic force between two test images with deforming objects is investigated, and the analysis of deforming is based on the distribution of the force on each discrete current element, because the deforming process may occur locally at each position in the image. Therefore, the force for each current element in the image is calculated and analyzed, based on which the details of deformation can be estimated.

Experiments have been designed to study the virtual force for two images with deformed objects. The experiment results for some test images with deforming objects are shown in Fig. 12 to Fig. 15. The test images are of the size $32 \times 32$. The original image and the deformed one are on the same plane when the virtual force distribution is calculated. To analyze the deforming process, the force on each current element in the original image applied by the deformed one is calculated, which is recorded in the force distribution. Each virtual force is calculated according to Equation (7). Here suppose the two images are on the same plane, and have the same coordinates.

Fig. 12 shows the case of deforming a rectangle to a square. Fig. 12(a) and Fig. 12(c) show the rectangle and the square image respectively. Fig. 12(b) and Fig. 12(d) show the significant edge lines as the virtual current in the two images. The virtual force distribution is shown in Fig. 12(e), in which the arrows show the discrete direction of the force on each current element in Fig. 12(b) applied by Fig. 12(d). Thus all the arrows in Fig. 12(e) show the border shape of Fig. 12(a). The gray area in Fig. 12(e) shows the shape of the square. It is obvious that the directions of the arrows in Fig. 12(e) clearly shows the deforming process from the rectangle to the square: according to the arrow directions, the rectangle's upper border goes up, and the lower border goes down; the left border goes right, and the right border goes left.

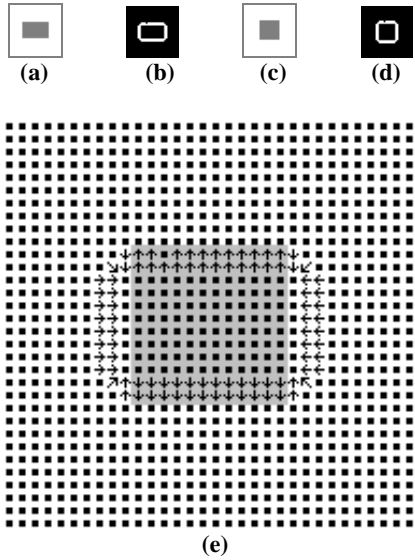

**Fig. 12 The force on each current element in the rectangle image (deforming to the square)**
**(a) the rectangle image**
**(b) the significant edge lines in (a)**
**(c) the square image**
**(d) the significant edge lines in (b)**
**(e) the force on each current element in (b) applied by the virtual currents in (d)**

Fig. 13 shows the case of deforming an irregular shape to a square. Fig. 13(a) and Fig. 13(c) show the irregular shape and the square image respectively. Fig. 13(b) and Fig. 13(d) show the significant edge lines as the virtual current in the two images. The virtual force distribution is shown in Fig. 13(e), in which the arrows show the discrete direction of the force on each current element in Fig. 13(b) applied by Fig. 13(d). Thus all the arrows in Fig. 13(e) show the border shape of Fig. 13(a). The gray area in Fig. 13(e) shows the shape of the square. It is obvious that the directions of the arrows in Fig. 13(e) clearly show the deforming process from the irregular shape to the square: according to the arrow directions, some parts of the shape's border shrink toward the center, and other parts expand outward.

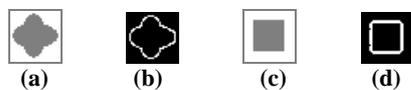

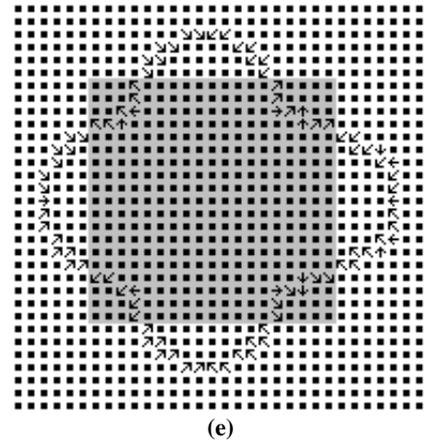

**Fig. 13 The force on each current element in the image of irregular shape (deforming to the square)**
**(a) the image of a irregular shape**
**(b) the significant edge lines in (a)**
**(c) the square image**
**(d) the significant edge lines in (b)**
**(e) the force on each current element in (b) applied by the virtual currents in (d)**

Another two examples of the circle are shown in Fig. 14 and Fig. 15. Fig. 14 shows the case of shrinking a larger circle to a smaller one. Fig. 14(a) and Fig. 14(c) show the two circles respectively. Fig. 14(b) and Fig. 14(d) show the significant edge lines as the virtual current in the two images. The virtual force distribution is shown in Fig. 14(e), in which the arrows show the discrete direction of the force on each current element in Fig. 14(b) applied by Fig. 14(d). Thus all the arrows in Fig. 14(e) show the border shape of Fig. 14(a). It is obvious that the directions of the arrows in Fig. 14(e) clearly show the shrinking process: all the arrows point toward the center, which corresponds to a shrinking process.

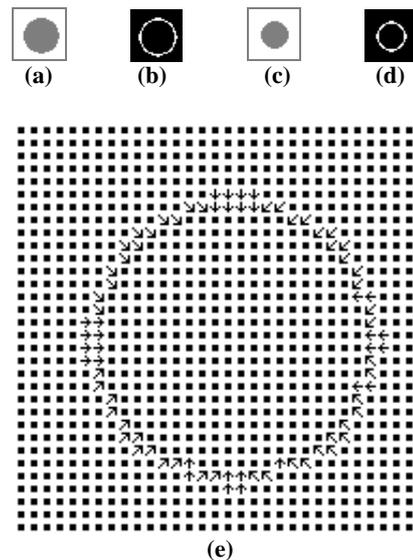

**Fig. 14 The force on each current element in the image of large circle (deforming to the small circle)**

(a) the image of large circle
(b) the significant edge lines in (a)
(c) the image of small circle
(d) the significant edge lines in (b)
(e) the force on each current element in (b) applied by the virtual currents in (d)

Fig. 15 shows the case of deforming an ellipse to a circle. Fig. 15(a) and Fig. 15(c) show the ellipse and the circle image respectively. Fig. 15(b) and Fig. 15(d) show the significant edge lines as the virtual current in the two images. The virtual force distribution is shown in Fig. 15(e), in which the arrows show the discrete direction of the force on each current element in Fig. 15(b) applied by Fig. 15(d). Thus all the arrows in Fig. 15(e) show the border shape of Fig. 15(a). It is obvious that the directions of the arrows in Fig. 15(e) clearly shows the deforming process: according to the arrow directions, the left and right ends of the ellipse shrink towards the center, while its upper and lower sides expand outwards.

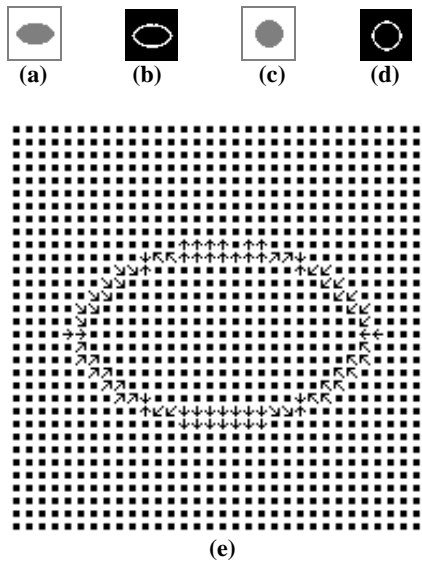

Fig. 15 The force on each current element in the ellipse image (deforming to the circle)
(a) the ellipse image
(b) the significant edge lines in (a)
(c) the circle image
(d) the significant edge lines in (b)
(e) the force on each current element in (b) applied by the virtual currents in (d)

## 5 Local Motion and Deformation Estimation in Video Sequences by Virtual Magnetic Force

The experiment results for test images indicate that the virtual force distribution can effectively represent the object's deforming process between two images. For practical use, in image sequence processing such as video analysis, the local change between adjacent frames is important, because the analysis of the deformations and motions of objects is mainly based on the change between adjacent frames. Moreover, the motion and deformation of an object often occur simultaneously for an object or region in practice video sequences. On the other hand, if there is no dramatic change between adjacent frames (which is commonly seen in modern video capture devices with sufficiently high frame rate of recording), the deformations or movements of objects is also relatively small, which is commonly seen in videos. The virtual electromagnetic interaction method is quite suitable to such situation. (Here the intensity of change between frames is related to the frame rate in image sequence recording. Fast frame rate in recording can capture more continuous details of changing.) The virtual force distribution can reflect the local changes between two related images. Such local changes can further be used in analysis of deformation or movement in image sequences. Experiments are carried out for real world image sequences, and some of the experiment results are shown in Fig. 16 to Fig. 19.

Face image processing has been widely used in identity recognition, emotion recognition, speech recognition, etc [22-26]. In audio-visual speech recognition, the analysis of lip movement is a key issue. Experiments have been carried out for some videos of speakers. Fig. 16 shows an example. Fig. 16(a) and Fig. 16(b) show two frames in a video sequence. The speaker closed the lips from frame1 to frame2. Fig. 16(c) and Fig. 16(d) show the significant edge lines extracted as the virtual currents of the two frames. Fig. 16(e) shows the virtual force distribution on the mouth border lines extracted from Fig. 16(c). In Fig. 16(e), according to the arrow directions, the upper mouth border moves down and the lower border moves up, which clearly indicates that a mouth-closing action occurs from frame1 to frame2.

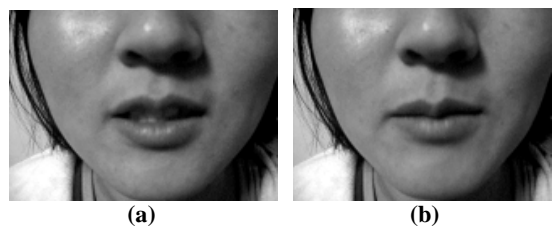

(a)　　　　　　　　(b)

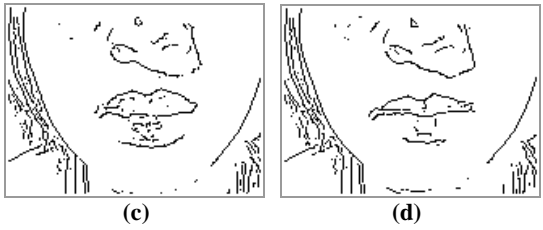

(c)　　　　　　　　(d)

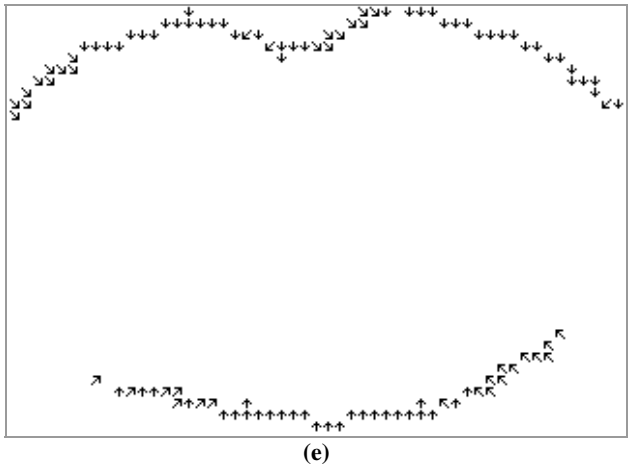

(e)

**Fig. 16 The forces on the mouth borders in the frame of the first speaker video sequence**
**(a) the 1st frame**
**(b) the 2nd frame**
**(c) the significant edge lines in (a)**
**(d) the significant edge lines in (b)**
**(e) the force on each current element on the mouth borders in (c) applied by the virtual currents in (d)**

Another example is shown in Fig. 17 for comparison, in which the speaker opened the mouth. Fig. 17(a) and Fig. 17(b) show the two frames, and their significant edge lines are shown in Fig. 17(c) and Fig. 17(d). Fig. 17(e) shows the virtual force distribution on the mouth border lines extracted from Fig. 17(c). It is obvious in Fig. 17(e) that the upper border of mouth moves up and the lower border moves down, which indicates a mouth opening action. These results can be used in further analysis and recognition tasks.

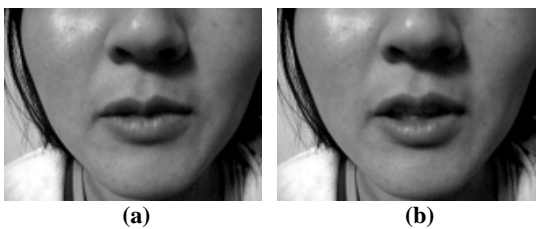

(a)　　　　　　　　(b)

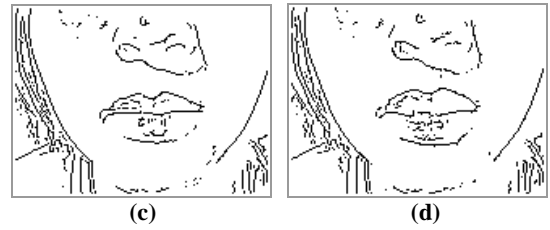

(c)　　　　　　　　(d)

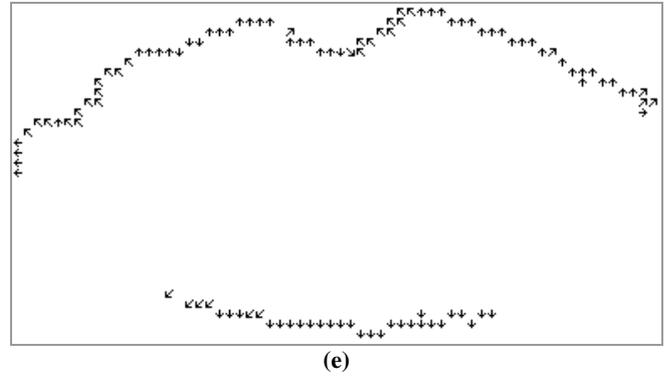

(e)

**Fig. 17 The forces on the mouth borders in the frame of the second speaker video sequence**
**(a) the 1st frame**
**(b) the 2nd frame**
**(c) the significant edge lines in (a)**
**(d) the significant edge lines in (b)**
**(e) the force on each current element on the mouth borders in (c) applied by the virtual currents in (d)**

Fig. 18 shows an example of a medical image sequence. The two frames are shown in Fig. 18(a) and Fig. 18(b), in which some areas deform. Fig. 18(c) and Fig. 18(d) show their significant edge lines. In the virtual force distribution, two local areas are investigated for analysis. The first one is on the middle left of the frame, which is highlighted in Fig. 18(e). The virtual force distribution in this area is shown in Fig. 18(f), in which an expanding process of the object area can be clearly seen according to the arrow directions.

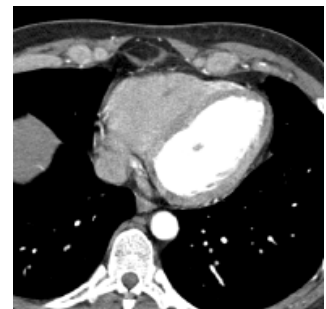

**(a) the 1st frame**

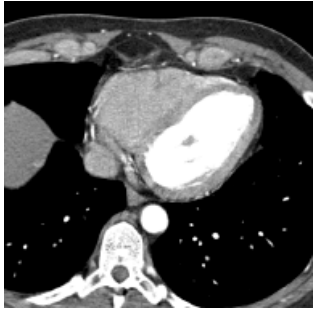

**(b) the 2nd frame**

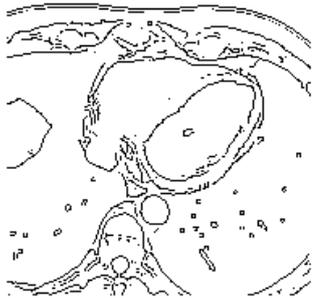

**(c) the significant edge lines of (a)**

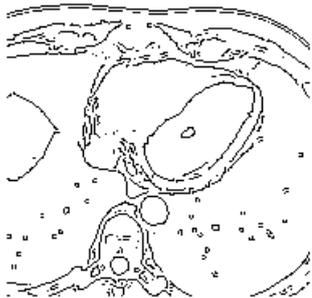

**(d) the significant edge lines of (b)**

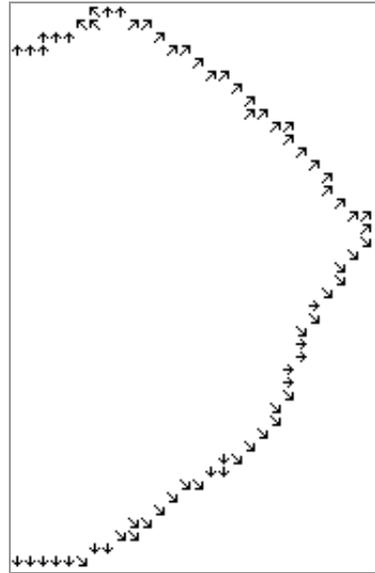

**(f) the forces on the current elements for the local area in (e)**

**Fig. 18 The forces on the current elements for a local area of a medical sequence frame**

The second local area is near the center of the frame, which is highlighted in Fig. 19(a). The virtual force distribution in this area is shown in Fig. 19(b), in which an expanding process of the object area can be clearly seen according to the arrow directions. Such results may be helpful in detailed analysis and diagnosis.

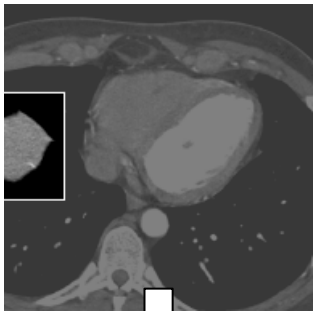
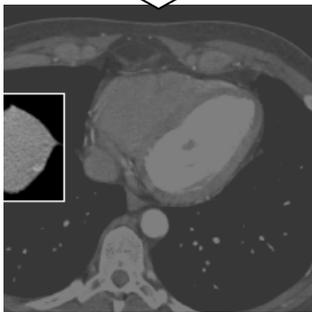

**(e) the highlighted local area in the two frames**

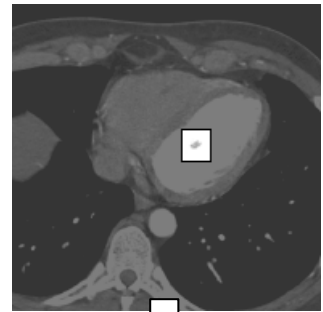
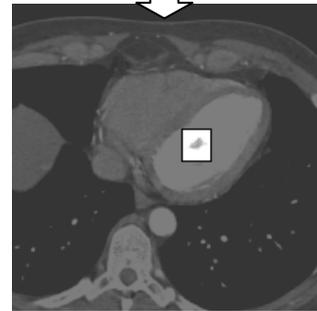

**(a) the highlighted local area in the two frames**

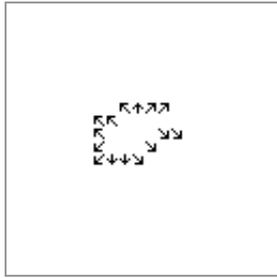

**(b) the forces on the current elements for the local area in (a)**
**Fig. 19 The forces on the current elements for another local area of a medical sequence frame**

Besides deforming, the virtual force distribution can also reveal other movements between two adjacent frames. Experiments are carried out on some videos of human actions. Fig. 20 and Fig. 21 show two examples of arm-raising action, in which the person raised the arms on the grassland. Fig. 20 shows the first example. The two frames are shown in Fig. 20(a) and Fig. 20(b), and the significant edge lines extracted are shown in Fig. 20(c) and Fig. 20(d). Fig. 20(e) shows the force distribution on the arms in Fig. 20(c) applied by Fig. 20(d). Almost all the arrows on the borders of the arm area go upwards, which indicates an arm-raising action.

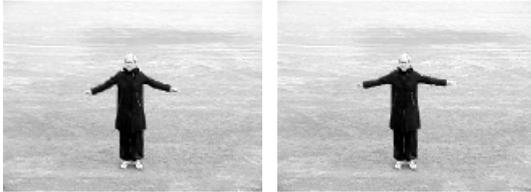

(a)          (b)

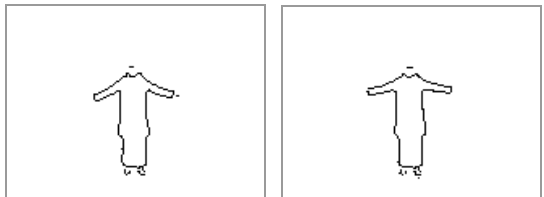

(c)          (d)

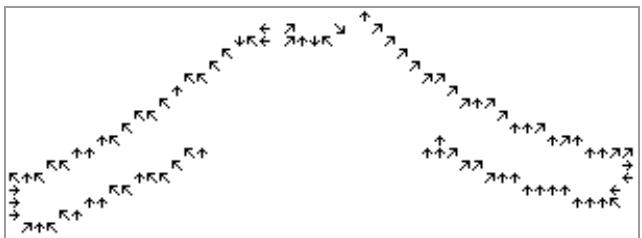

(e)

**Fig. 20 The forces on the arm borders in the frame of the first arm-raising video sequence**
(a) the 1st frame
(b) the 2nd frame
(c) the significant edge lines in (a)
(d) the significant edge lines in (b)
(e) the force on each current element on the arm borders in (c) applied by the virtual currents in (d)

Similar result can be found in Fig. 21. The two frames are shown in Fig. 21(a) and Fig. 21(b), and the significant edge lines extracted are shown in Fig. 21(c) and Fig. 21(d). The force distribution on the arms in Fig. 21(e) also indicates an arm-raising action. Such results may be useful in automatic human action recognition in video sequence.

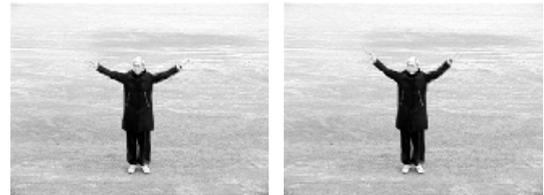

(a)          (b)

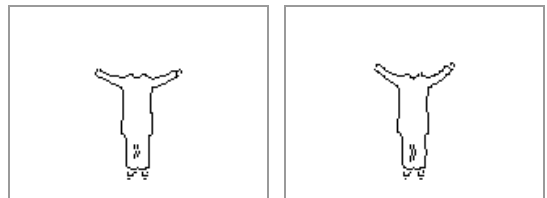

(c)          (d)

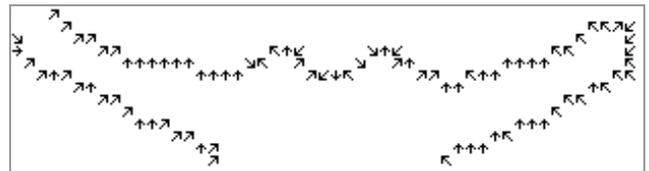

(e)

**Fig. 21 The forces on the arm borders in the frame of the second arm-raising video sequence**
(a) the 1st frame
(b) the 2nd frame
(c) the significant edge lines in (a)
(d) the significant edge lines in (b)
(e) the force on each current element on the arm borders in (c) applied by the virtual currents in (d)

Two examples of hand actions are shown in Fig. 22 and Fig. 23. The two frames in Fig. 22(a) and Fig. 22(b) show a clenching action. The significant edge lines are shown in Fig. 22(c) and Fig. 22(d). For demonstration, the virtual force distribution on the thumb and the index finger is shown in Fig. 22(e). According to the arrow directions, the upper part of the index finger moves down left, and its lower part moves right, which is a bending movement towards the thumb. On the other hand, the thumb draws close to the index finger. The virtual force distribution here corresponds to a typical clenching action.

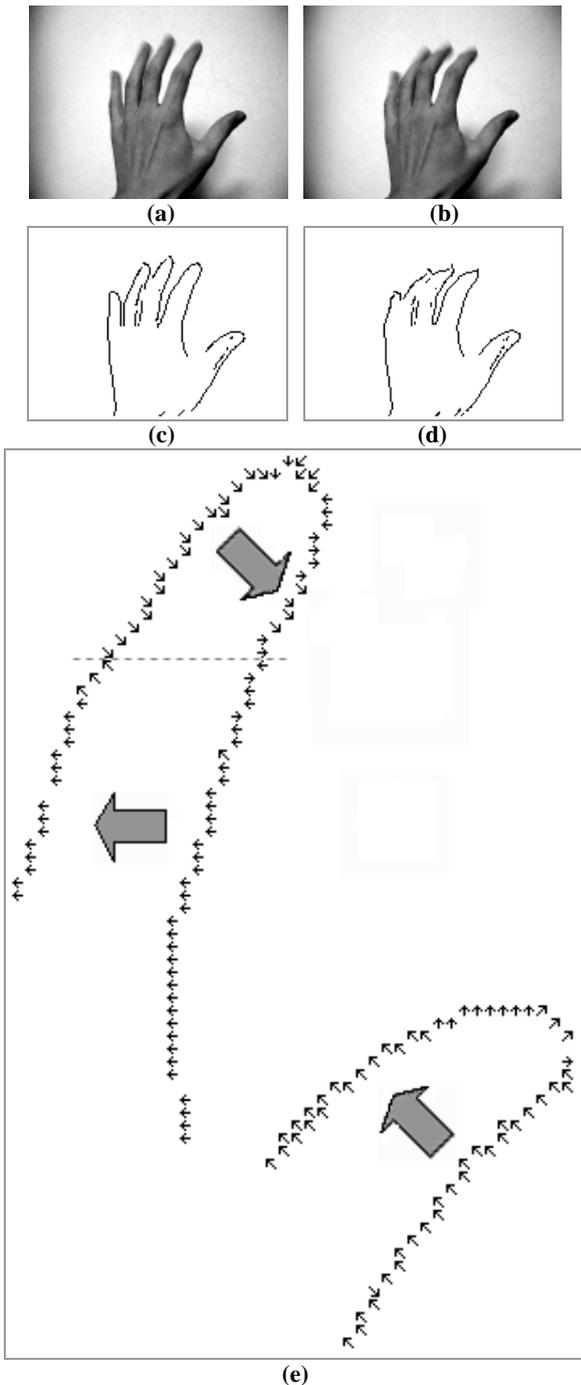

**Fig. 22 The forces on the finger borders in the frame of the hand-clenching video sequence**
(a) the 1st frame
(b) the 2nd frame
(c) the significant edge lines in (a)
(d) the significant edge lines in (b)
(e) the force on each current element on the borders of the thumb and index finger in (c) applied by the virtual currents in (d)

The two frames in Fig 23(a) and Fig. 23(b) show a hand-waving action. The significant edge lines are shown in Fig. 23(c) and Fig. 23(d). For demonstration, the force distribution in a local area of the ring finger and the small finger is shown here. The local area is highlighted in Fig. 23(e), and the force distribution on the fingers for that local area is shown in Fig. 23(f). In Fig. 23(f), almost all the arrows accord with a clockwise rotating, which correspond to a hand-waving action.

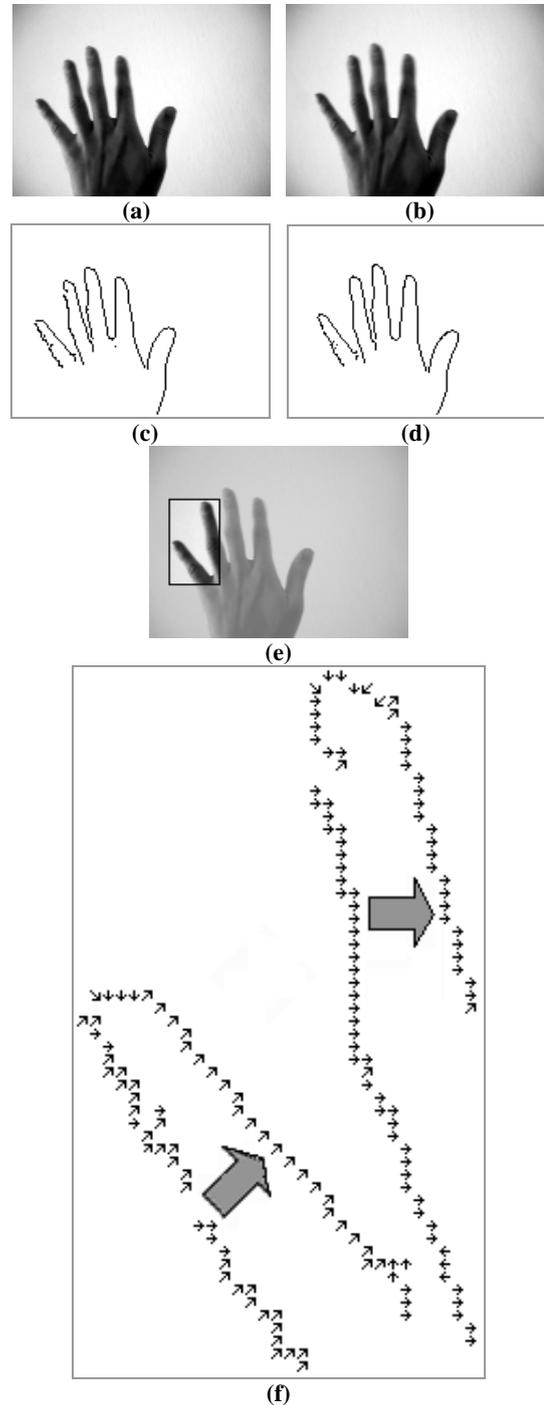

**Fig. 23 The forces on the finger borders for a local area in the frame of the hand-waving video sequence**
(a) the 1st frame
(b) the 2nd frame
(c) the significant edge lines in (a)
(d) the significant edge lines in (b)
(e) the highlighted local area
(f) the force on each current element in the highlighted local area in (e)

Based on the above experiments and analysis, it is indicated that the virtual force distribution between adjacent frames in image sequences can reveal the deformation and movements of objects, which can be used in further analysis. One important thing to note here is: the virtual force distribution for two frames directly shows the local change between the frames; if an object in the frames deforms and moves simultaneously, the virtual force distribution shows the composed result of the simultaneous transformations of that object.

# 6 Conclusion and Discussion

The electromagnetic interaction is a basic and important phenomenon in physics. The current-carrying wires generate magnetic field. On the other hand, the magnetic field applies force on other currents in it. The interesting phenomena of electromagnetic interaction can generate inspiring physical effects, which may be exploited in design novel image processing algorithms.

A macroscopic current in the wire is based on the microscopic movement of electrons. The virtual current proposed in the paper is a kind of reasonable imitation of physical current in the image space, which is the basis of the virtual electromagnetic interaction method proposed. In this paper, the virtual current in images is proposed based on the main edge lines. The significant edge lines are extracted from the edge current vectors by a Canny-like operation as the image's structure representation, based on which the virtual edge current is defined. The virtual interaction between the significant edge currents is studied by imitating the electro-magnetic interaction between the current-carrying wires.

The virtual force distribution consists of the virtual force on each current element in one image applied by all the current elements in the other image. Because the magnetic force has the local property that the force on a current element in one image is mainly determined by the adjacent current elements in the other image (suppose the two images are on the same plane, and coincide in position), the virtual magnetic force can reflect the local changes of edge between the two images. Therefore, the above virtual force distribution can represent the local change of position or shape of objects in image sequences, which provides the basis for further analysis. Future work will further analyze the motion and deforming, and also behavior recognition in image sequences based on the output of the proposed method in this paper.

The experimental results of the proposed method indicate that it can provide useful clues for further image sequence analysis and recognition. For specific tasks such as vehicle motion analysis or human action recognition, a further step is still needed to automatically integrated the related local edge changes into recognition results at a more global or abstract level (such as abstract description or classification of motions or actions). In another word, the proposed method here can output the local changes of significant edge lines, and it is our further research work in future to reasonably integrate related local change to global analysis results which can be directly used for automatic recognition (which will also be more comprehensible to us). On the other hand, the image sequences processed in this paper are of grayscale. For the color image sequence, the proposed method can be extended if a reasonable definition of "color gradient" is given, which is also a research topic to be investigated in future.

*References:*
[1] Antonios Oikonomopoulos, Ioannis Patras, Maja Pantic, Spatiotemporal localization and categorization of human actions in unsegmented image sequences, IEEE Transactions on Image Processing, Vol. 20, No. 4, 2011, pp. 1126-1140.
[2] Kanglin Chen, Dirk A. Lorenz, Image sequence interpolation based on optical flow, segmentation, and optimal control, IEEE Transactions on Image Processing, Vol. 21, No. 3, 2012, pp. 1020-1030.
[3] Iulian Udroiu, Ioan Tache, Nicoleta Angelescu, Ion Caciula, Methods of measure and analyse of video quality of the image, WSEAS Transactions on Signal Processing, Vol. 5, No. 8, 2009, pp. 283-292.
[4] Radu Dobrescu, Matei Dobrescu, Dan Popescu, Parallel image and video processing on distributed computer systems, WSEAS Transactions on Signal Processing, Vol. 6, No. 3, 2010, pp. 123-132.
[5] X. D. Zhuang, N. E. Mastorakis,The Relative Potential Field as a Novel Physics-Inspired Method for Image Analysis, WSEAS Transactions on Computers, Issue 10, Volume 9, 2010, pp. 1086-1097.
[6] Mark S. Nixon, Xin U. Liu, Cem Direkoglu, David J. Hurley, On using physical analogies for feature and shape extraction in computer vision, Computer Journal, Vol. 54, No. 1, 2011, pp. 11-25.
[7] D.J. Hurley, M.S. Nixon, J.N. Carter, A new force field transform for ear and face recognition,

IEEE International Conference on Image Processing, Vol. 1, 2000, pp. 25-28.
[8] David J. Hurley, Mark S. Nixon, John N. Carter, Force field feature extraction for ear biometrics, Computer Vision and Image Understanding, Vol. 98, No. 3, 2005, pp. 491-512.
[9] David J. Hurley, Mark S. Nixon, John N. Carter, Force field energy functionals for image feature extraction, Image and Vision Computing, Vol. 20, No. 5-6, 2002, pp. 311-317
[10] Xin U Liu, Mark S Nixon, Water Flow Based Complex Feature Extraction. Advanced Concepts for Intelligent Vision Systems, Lecture Notes in Computer Science, 2006. pp. 833-845.
[11] Xin U Liu, Mark S Nixon, Medical Image Segmentation by Water Flow, in Proceedings of Medical Image Understanding and Analysis, MIUA 2007.
[12] Xin U Liu, Mark S Nixon, Water flow based vessel detection in retinal images, Proceedings of IET International Conference on Visual Information Engineering 2006, 2006, pp. 345-350.
[13] Xin U Liu, Mark S Nixon, Image and volume segmentation by water flow, Third International Symposium on Proceedings of Advances in Visual Computing, ISVC 2007, 2007, pp. 62-74.
[14] P. Hammond, Electromagnetism for Engineers: An Introductory Course, Oxford University Press, USA, forth edition, 1997.
[15] I. S. Grant and W. R. Phillips, Electromagnetism, John Wiley & Sons, second edition, 1990.
[16] Terence W. Barrett, Topological foundations of electromagnetism, World Scientific series in contemporary chemical physics, Vol. 26, World Scientific, 2008.
[17] Minoru Fujimoto, Physics of classical electromagnetism, Springer, 2007.
[18] J. Canny, A Computational Approach To Edge Detection, IEEE Trans. Pattern Analysis and Machine Intelligence, 8(6), pp. 679–698, 1986.
[19] R. Deriche, Using Canny's criteria to derive a recursively implemented optimal edge detector, Int. J. Computer Vision, Vol. 1, pp. 167–187, 1987.
[20] Barbara Zitova, Jan Flusser: Image registration methods: a survey. Image Vision Comput. 21(11): 977-1000 (2003).
[21] Tim McInerney, Demetri Terzopoulos, Deformable models in medical image analysis: a survey, Medical Image Analysis, Volume 1, Issue 2, June 1996, pp. 91-108.
[22] Ming-hsuan Yang, David J. Kriegman, Narendra Ahuja, Detecting Faces in Images: A Survey, IEEE Transactions on Pattern Analysis and Machine Intelligence, vol. 24, no. 1, pp. 34-58, 2002.
[23] W. Zhao, R. Chellappa, A. Rosenfeld, P.J. Phillips, Face Recognition: A Literature Survey, ACM Computing Surveys, 2003, pp. 399-458.
[24] Jongju Shin , Jin Lee , Daijin Kim, Real-time lip reading system for isolated Korean word recognition, Pattern Recognition, Volume 44, Issue 3, March 2011, Pages 559-571.
[25] Piotr Dalka, Andrzej Czyzewski, Human-Computer Interaction Based on Visual Lip Movement and Gesture Recognition, International Journal of Computer Science and Applications, Vol. 7 No. 3, pp. 124 - 139, 2010.
[26] Issues in Visual and Audio-Visual Speech Processing, G. Bailly, E. Vatikiotis-Bateson, and P. Perrier, MIT Press, 2004